\documentclass[conference,10pt]{IEEEtran}
\usepackage{amsmath,graphicx}

\usepackage[T1]{fontenc}
\usepackage{graphicx}
\usepackage{textcomp}
\usepackage{xcolor}
\usepackage{amssymb,amsmath,amsfonts,amsthm}
\usepackage{mathrsfs}
\usepackage{cite}
\usepackage{array}
\usepackage{tabularx}
\usepackage{color}
\usepackage{subcaption}
\usepackage{mathtools}
\usepackage{tikz,pgf}
\usetikzlibrary{calc,positioning,mindmap,trees,decorations.pathreplacing}
\usepackage{graphicx}
\usepackage{bbm}
\usepackage{bm}
\usepackage[utf8]{inputenc}
\usepackage{algorithm}
\usepackage{algpseudocode}
\usepackage{stfloats}

\newcommand{\matr}[1]{\mathbf{#1}}
\newcommand{\vect}[1]{\mathbf{#1}}

\usepackage[colorinlistoftodos,prependcaption,backgroundcolor=black!5!white,bordercolor=red]{todonotes}

\begin{document}
	\title{Decentralized Learning over Wireless Networks with Broadcast-Based Subgraph Sampling}

	\author{Daniel P\'erez Herrera, Zheng Chen and Erik G. Larsson\\
			Department of Electrical Engineering (ISY), Link\"{o}ping University, Sweden. \\
			Email:\{daniel.perez.herrera, zheng.chen, erik.g.larsson\}@liu.se\thanks{This work was supported by Zenith, ELLIIT, the KAW foundation, and the Swedish Research Council (VR).}}

	\maketitle
	
	\begin{abstract}
		This work centers on the communication aspects of decentralized learning over wireless networks, using consensus-based decentralized stochastic gradient descent (D-SGD). Considering the actual communication cost or delay caused by in-network information exchange in an iterative process, our goal is to achieve fast convergence of the algorithm measured by improvement per transmission slot. 
		We propose BASS, an efficient communication framework for D-SGD over wireless networks with broadcast transmission and probabilistic subgraph sampling. 
		In each iteration, we activate multiple subsets of non-interfering nodes to broadcast model updates to their neighbors. These subsets are randomly activated over time, with probabilities reflecting their importance in network connectivity and subject to a communication cost constraint (e.g., the average number of transmission slots per iteration). During the consensus update step, only bi-directional links are effectively preserved to maintain communication symmetry. In comparison to existing link-based scheduling methods, the inherent broadcasting nature of wireless channels offers intrinsic advantages in speeding up convergence of decentralized learning by creating more communicated links with the same number of transmission slots.
	\end{abstract}
	
	
	\section{Introduction}
	Collaborative training of Machine Learning (ML) models across networked agents allows users optimize a common ML model over the union of their local data without directly sharing the data \cite{kairouz2021advances}. The prevailing research in collaborative machine learning has concentrated on server-based federated learning \cite{mcmahan2017communication}. In the meanwhile, decentralized learning with locally connected agents has gained significant attention and interest recently due to its scalability and resistance to single-node failure \cite{koloskova2020unified}. 
	
	Consensus-based decentralized stochastic gradient descent (D-SGD) is a widely used first-order optimization method for decentralized learning \cite{swenson2020distributed}.
	The plain version of D-SGD follows an iterative procedure where in every iteration each agent updates its model by combining the models received from its neighbors and its own stochastic gradient.
	The convergence of D-SGD has been widely studied in the literature \cite{wang2021cooperative, lian2017can, jakovetic2018convergence}. In the predominant body of literature, the convergence speed of D-SGD algorithm is characterized by the level of error reduction per iteration, while the actual runtime (or communication costs) in every iteration is neglected. 
	As pointed out in \cite{wang2022matcha}, there is a fundamental trade-off between the runtime per iteration and the error-versus-iterations convergence. This aspect is even more profound when implementing D-SGD over wireless networks, where access control and communication coordination is needed to reduce information loss caused by packet collision. The specific choice of access control protocol determines the amount of communication costs (e.g., transmission slots) in every consensus updating iteration. In some cases, introducing partial communication where a subset of nodes or links are activated in every iteration may accelerate the convergence of D-SGD in terms of error reduction per communication cost.
	
	Communication-efficient decentralized learning has been studied in many existing works \cite{koloskova2019decentralized, tang2018communication}, where the main focus is on the impact of model compression to reduce the amount of data transmitted over every link. 
	Others works \cite{wang2019adaptive, tran2019federated} have explored graph sparsification techniques for improving convergence speed with reduced communication frequency. 
	Following this line, more recent works such as \cite{wang2022matcha} and \cite{chiu2023laplacian} have pointed out an important message that justifies the use of graph sparsification: ``\textit{not all links are equally important in a graph}''. Fast convergence can be achieved by allowing more important links to communicate more often, while the importance of a link is characterized by its impact on the connectivity of the communication graph.
	
	Another significant feature of wireless networks is their broadcasting nature. 
	With one broadcast transmission, a node can reach all its neighbors, increasing the level of information dissemination in the network.
	In our work, the main motivation is that ``\textit{not all  nodes are equally important in a graph}''. We propose a broadcast-based node scheduling scheme for D-SGD over wireless networks, wherein in each iteration, we sample subgraphs of the base topology with probabilities related to the importance of the nodes contained in the subgraphs.
	To address the asymmetric information flow associated with broadcast communication, we also present a general method for generating symmetric Laplacian matrices from directed subgraphs of a given base topology. Additionally, we optimize the weight matrix to accelerate convergence. As compared to similar designs with link-based scheduling \cite{wang2022matcha}, our proposed method BASS shows significant performance gain under the same communication cost, especially when the network topology contains densely connected local structures.

	\section{Preliminaries on Decentralized Learning}
	We consider a network of $N$ agents where the communication topology is represented by an undirected and connected graph $\mathcal{G}=(\mathcal{N},\mathcal{E})$ with node/agent set $\mathcal{N}=\{1,\ldots, N\}$ and edge/link set $\mathcal{E}\subseteq\mathcal{N}\times\mathcal{N}$. Each node $i$ can only communicate with its set of neighbors, denoted as $\mathcal{N}_i = \{ j|\{i,j\}\in \mathcal{E}\}$. 
	The adjacency matrix of the graph is defined as $\matr{A}\in \mathbb{R}^{N\times N}$. In particular, it will have elements $a_{ij} = 1$ if $\{i,j\}\in \mathcal{E}$ and  $a_{ij} = 0$ otherwise. The Laplacian matrix is defined as $\matr{L} = \text{diag}(d_1, \ldots, d_N) - \matr{A}$, where $d_i=|\mathcal{N}_i|$ is the degree of node $i\in\mathcal{N}$.
	
	Each node has a locally available set of training data samples. The objective is for all nodes to collaboratively train a machine learning model in a decentralized setting through local computation and information exchange among locally connected neighbors. 
	
	\subsection{Collaborative Machine Learning in Decentralized Setting}
	\label{requirements_W}
	
    Let $\vect{x}\in\mathbb{R}^d$ represent the learning model. The objective of collaborative training can be written as the following optimization problem 
	\begin{equation}
		\min_{\vect{x}\in\mathbb{R}^d} \frac{1}{N}\sum_{i=1}^N F_i(\vect{x}),
	\end{equation}
	where each $F_i:\mathbb{R}^d\rightarrow\mathbb{R}$ defines the local objective function of agent $i$. For model training, the local objective $F_i$ can be defined as the expected local risk/loss function
	\begin{equation}
		F_i(\vect{x}):=\mathbb{E}_{s\sim\mathcal{D}_i}f(\vect{x};s),
	\end{equation}
	where $\mathcal{D}_i$ denotes the local data distribution at agent $i$, and $f(\vect{x};s)$ is the loss function of the learning model $\vect{x}$ for sample $s$.
	
	To perform collaborative model training in decentralized systems, we use consensus-based decentralized SGD (or D-SGD) \cite{lian2017can}, that relies on local gradient updating and consensus-based averaging. Every iteration of the algorithm includes the following steps:
	\begin{enumerate}
		\item \textbf{Stochastic gradient update:} Each node $i$ computes its stochastic gradient vector $\vect{g}_i^{(t)}$ and updates its local model according to 
		\begin{equation}
		\vect{x}_i^{(t+\frac{1}{2})} = \vect{x}_i^{(t)}-\gamma \vect{g}_i^{(t)},
		\end{equation} 
		where $\gamma$ denotes the learning rate and $t$ is the iteration index. The stochastic gradient of each node is computed based on a mini-batch of samples drawn from its local data distribution $\mathcal{D}_i$, i.e., $\vect{g}_i^{(t)}=\frac{1}{|\xi_i|}\sum_{s\in\xi_i}\nabla f(\vect{x}_i;s)$ with $\xi_i\sim \mathcal{D}_i$.
		\item \textbf{Consensus averaging:} Each node shares its local model with its neighbors and obtains an averaged model as
		\begin{equation}
			\vect{x}_i^{(t+1)}=\sum_{j=1}^N w_{ij}\vect{x}_j^{(t+\frac{1}{2})},
		\end{equation}  
		where $w_{ij}$ is the $(i,j)$-th element of a matrix $\matr{W}\in \mathbb{R}^{N\times N}$ that is used to encode how much nodes affect each other, and the graph it induces is consistent with the base topology, meaning that if $a_{ij}=0$, then $w_{ij}=0$.
	\end{enumerate}

	This matrix $\matr{W}$ is known as the mixing matrix \cite{wang2022matcha}. For fixed topologies, the convergence of consensus averaging (step 2) is guaranteed when it is symmetric ($\matr{W}^{T} = \matr{W}$), with each row/column summing up to one ($\matr{W}\vect{u} = \vect{u}$, $\vect{u}^T\matr{W} = \vect{u}^T$, where $\vect{u}$ is a column vector of all-ones), and $\rho(\matr{W}-\matr{J}) < 1$, where $\rho(\cdot)$ denotes the spectral radius of a matrix, and $\matr{J}=\vect{u}\vect{u}^T/N$. 

	D-SGD converges in the mean-square sense if in addition to these conditions on $\matr{W}$, the local objective functions $F_i(\vect{x})$ are differentiable with Lipschitz gradients, and the variances of the stochastic gradients are bounded \cite{lian2017can},\cite{chiu2023laplacian}. 

	One common choice of constructing the mixing matrix is to use Laplacian-based weight design:
	\begin{equation}
		\matr{W} = \matr{I} - \epsilon\matr{L}.
		\label{definition_W}
	\end{equation}
	with parameter $\epsilon>0$ \cite{olfati2007consensus}. This design assigns equal weights to all edges in the graph.
	Note that $\epsilon$ has to be chosen such that $\rho(\matr{W}-\matr{J})<1$.
	With fixed topology, the convergence rate and error bound of D-SGD are directly associated to $1- \rho(\matr{W}-\matr{J})$, which is the spectral gap of $\matr{W}$ \cite{lian2017can,neglia2020decentralized}. The parameter $\epsilon$ can be adjusted to obtain smaller $\rho(\matr{W}-\matr{J})$ and thus achieve better convergence performance.
	
	\section{D-SGD over Wireless Networks}
	For decentralized training, the information exchange among locally connected agents is the key factor for the convergence of the learning algorithm.
	One distinctive feature of wireless networks is that the communication channels are inherently broadcast channels. 
	At the cost of one transmission, a node can share its current local model to all its neighbors. 
	However, if all nodes broadcast at the same time, the information reception will fail due to packet collisions. In general, there are two types of approaches for coordinating multiple access communication in wireless networks: 
	\begin{itemize}
		\item random access, i.e., nodes make random decisions to access the channel and transmit their information. This policy is easy to implement and requires no centralized coordination, but it is prone to collisions;
		\item orthogonal division of resources, i.e., nodes are divided into collision-free sets and different transmission slots are allocated to different sets to avoid collision. 
	\end{itemize}
	
	In this work, we focus on the collision-free approach. For any pair of nodes $(i,j)$, the information transmission from node $i$ to $j$ is successful only if all other neighbors of $j$ are not transmitting. This condition imposes constraints on the set of links that can be simultaneously active.
	
\begin{figure}[t!]
		\centering
		\includegraphics[scale=0.35]{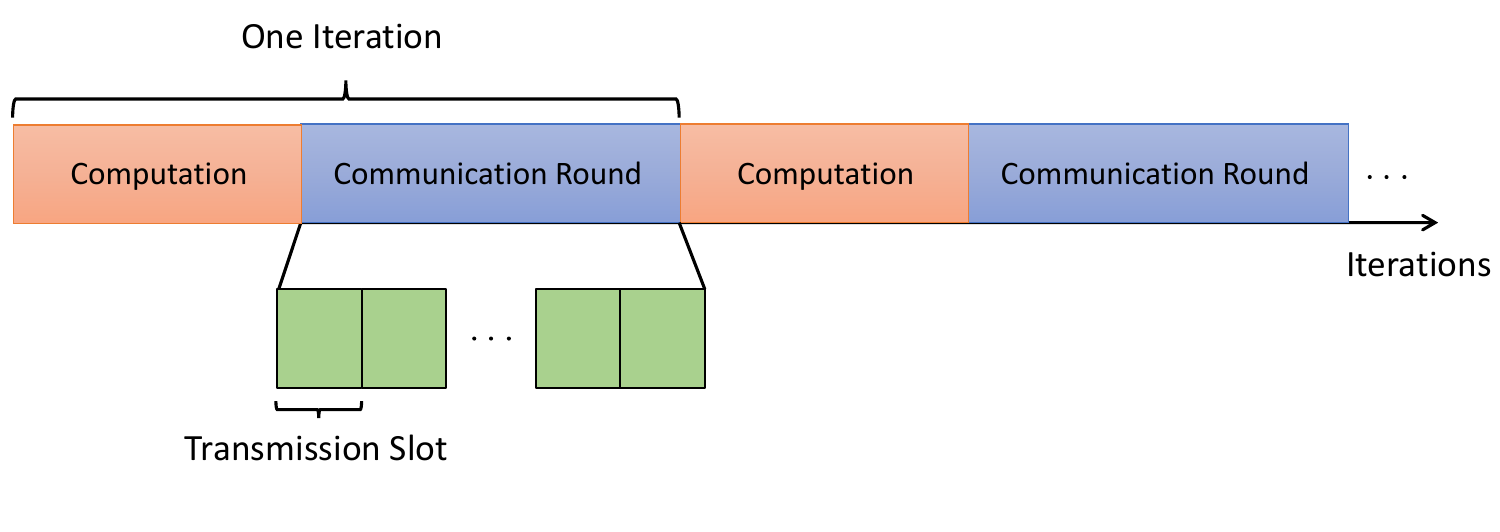}
		\caption{Communication Round vs. Transmission Slots}
		\label{transmission_slots}
	\end{figure}

	We consider a communication protocol wherein one communication round (one iteration of the D-SGD algorithm) is divided into multiple transmission slots, as illustrated in Fig. \ref{transmission_slots}. Within each transmission slot, a subset of non-interfering nodes are scheduled for broadcasting model updates to their neighbors. Furthermore, we consider the case of partial communication, meaning that not all subsets will be activated in every round. In this case, the number of transmission slots per round is generally smaller than the number of collision-free subsets.
	
	In Section \ref{partitioning}, we propose a subgraph partitioning mechanism to design these collision-free subsets.
	In Section \ref{probabilistic_design}, we present a probabilistic approach to make the scheduling decisions, assigning higher activation probabilities to the subsets with more ``important'' nodes/links.
	Partial communication with broadcast transmission introduces additional challenges, such as a mixing matrix asymmetry, which will be addressed in Section \ref{symm_lap_construction}. 
	Finally, we optimize the mixing matrix for faster convergence in Section \ref{mixing_opt}, and compare our method with other similar works in Section \ref{comparison_and_discussion}. Our proposed design is named as BASS, which stands for BroAdcast-based Subgraph Sampling.
	
	\subsection{Broadcast-based Subgraph Partition}
	\label{partitioning}
	
 	In wireless networks, information transmission and reception is always directed. To reflect the direction of communication, instead of considering an undirected graph $\mathcal{G}$ for the communication topology, we view it as a directed graph $\mathcal{G}^d=(\mathcal{V},\mathcal{E}^d)$ with bi-directional links between each pair of connected nodes. 
 
 	With broadcast-based transmission, each broadcasting node creates a local ``star'' with directed links towards its neighbors. Considering the collision-free condition of simultaneously transmitting nodes, we divide the base topology into disjoint subsets of nodes and their outgoing directed links $\mathcal{S}_k = \{\mathcal{V}_k,\mathcal{E}_k\}, k\in\{1,...,q\}$, where $q\leq N$. The elements in $\mathcal{V}_k$ are nodes that can transmit simultaneously without collision, i.e. $i,j\in\mathcal{V}_k$ if $\mathcal{N}_i\cap\mathcal{N}_j = \emptyset$, and $\mathcal{E}_k$ contains the outgoing links of all nodes in $\mathcal{V}_k$. 
	The partition satisfies $\cup_{k=1}^q\mathcal{V}_k = \mathcal{V}, \cup_{k=1}^q\mathcal{E}_k = \mathcal{E}^d$, $\mathcal{V}_k\cap\mathcal{V}_l=\emptyset$, and $\mathcal{E}_k\cap\mathcal{E}_l=\emptyset,\forall k\neq l$. 
	Note that these subsets are technically not subgraphs of $\mathcal{G}$, since they do not contain the nodes being pointed by the directed links.
	
	Finding an optimal partition (in the sense of finding a partition with the smallest number of subsets $q$) is a combinatorial optimization problem that, in general, is NP-hard. 
	Some heuristic algorithms can be used, such as the greedy vertex-colouring algorithm \cite{bollobas1998modern}, that assign different colors to connected nodes.
	
	To address the constraint in our setting, that two nodes cannot transmit simultaneously (be in the same subset) if they share a common neighbor, we follow a similar approach as in \cite{xing2021federated}. We run the coloring algorithm over an auxiliary graph $\mathcal{G}^a=(\mathcal{V},\mathcal{E}^a)$ such that $\mathcal{E}^a\supseteq\mathcal{E}$ includes the edges in $\mathcal{E}$ and an additional edge between each pair of nodes sharing at least one common neighbor.
	Then, nodes with the same color, and their outgoing links belong to the same subset, namely the \textit{collision-free subset}, and the number of subsets $q$ is equal to the number of colors found. 
	In Fig. \ref{Link_removal}(b) we show an example of dividing the base topology into multiple collision-free subsets, where each subset is marked with a different color.

	\subsection{Partial Communication with Probabilistic Scheduling}
	\label{probabilistic_design}
	When a collision-free subset is activated, all nodes within the subset can broadcast to their neighbors at the cost of one transmission slot. If all subsets are activated in one round, which represents the \textit{full communication} case, the total communication cost per round will be $q$ transmission slots.
	This cost can be reduced if partial communication with graph sparsification is considered, where only a fraction of nodes/links are scheduled in every round \cite{wang2022matcha, chiu2023laplacian}. 
	With partial communication, per-round convergence might be slowed down, but it can accelerate convergence in terms of actual runtime of the algorithm, especially under a limited communication budget. 
	Similar to \cite{wang2022matcha} and \cite{chiu2023laplacian}, we adopt a probabilistic scheduling/sampling approach, leading to a random number of activated subsets where each subset contains sparsified links from the base topology. With BASS, by taking into account the effect of broadcast transmission, more links can be activated under the same communication constraint.
	
	In every iteration of the D-SGD algorithm, each subset is scheduled with a certain probability.
	Let  $s_i(t)$ represent the scheduling decision of subset $\mathcal{S}_i$ in iteration $t$, which is equal to $1$ if $\mathcal{S}_i$ is activated, and $0$ otherwise.
	The activated communication topology $\mathcal{G}(t)=\{\mathcal{V}(t),\mathcal{E}(t)\}$ is defined as the union of all active subsets in iteration $t$, i.e, $\mathcal{G}(t) = \cup_{i=1}^qs_i(t)\mathcal{S}_i$. Note that $\mathcal{G}(t)$ is in general a directed graph, and only when $s_i(t)=1$ for all $i$, $\mathcal{G}(t) = \mathcal{G}^d$.
	
	\subsection{Constructing Symmetric Laplacian from Directed Graphs}
	\label{symm_lap_construction}

	Broadcast-based scheduling can be viewed as sampling columns of the adjacency matrix of the base topology, keeping them as they are, and masking the rest as zero. 
	The resulting matrix will represent a directed graph with edges pointing from the activated nodes (columns) towards their neighbors.
	
	If we sample columns of the adjacency matrix, and the rows with the same indices as the sampled columns (e.g. if we sample column $i$ we also sample row $i$), the resulting matrix will represent a subgraph with bidirectional links, removing the links being used in a single direction. 
	This ensures the symmetry of the resulting adjacency matrix and Laplacian, thus leading to symmetric mixing matrix as desired. From the consensus averaging perspective, it means that a node will pull an update from a neighbor only when it has pushed an update to that neighbor.
	
\begin{figure}[t!]
		\centering
		\includegraphics[width=\columnwidth]{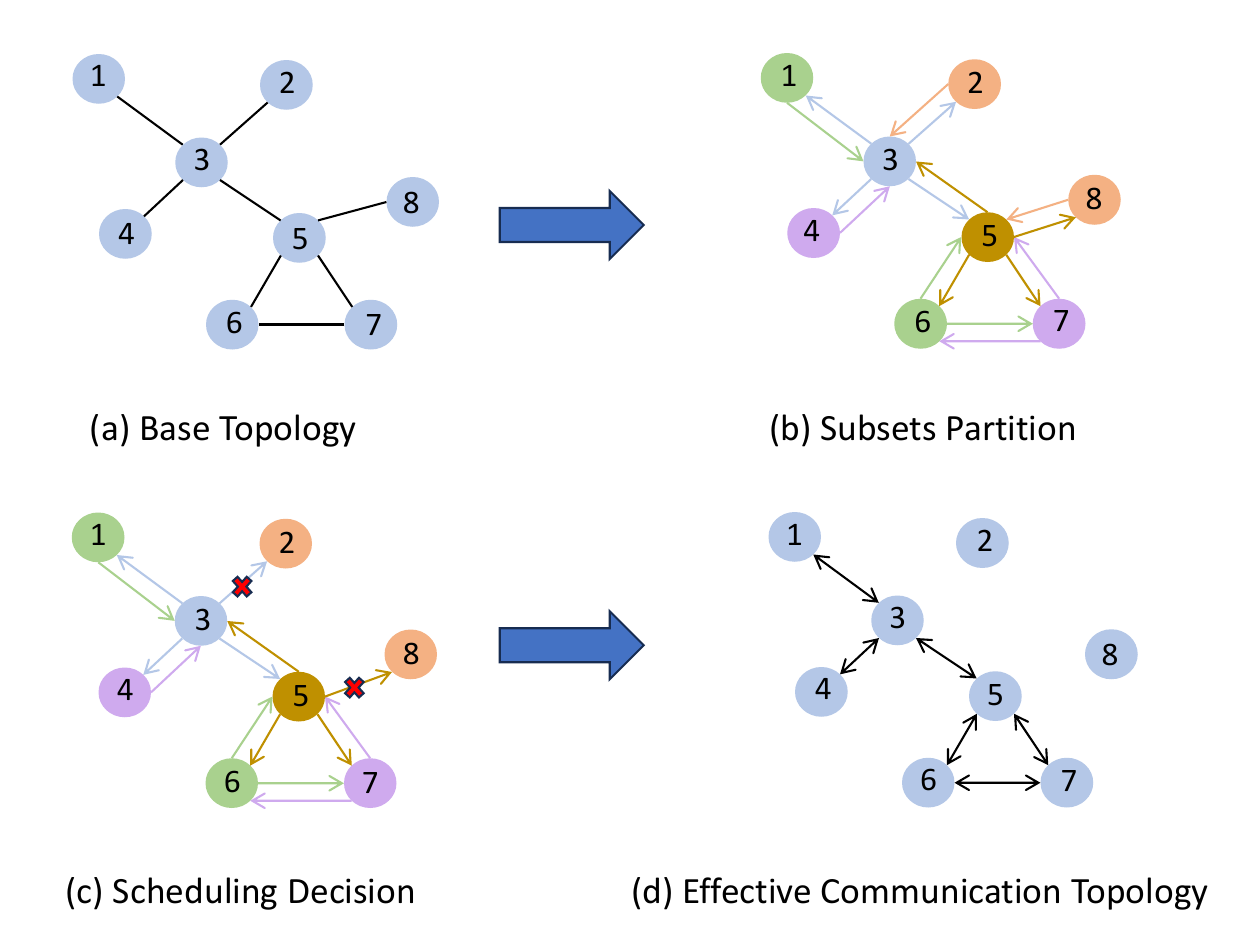}
		\caption{(a) Initial undirected base topology. (b) Subsets of nodes and their outgoing directed links. (c) Scheduling decision with links that are not used in both directions removed. (d) Effective topology with bi-directional links.}
		\label{Link_removal}
	\end{figure}

	Mathematically, the construction of a symmetric adjacency matrix by sampling columns and rows of $\matr{A}$ can be described as
	\begin{equation}
		\tilde{\matr{A}}(t)=\matr{Q}(t)\matr{A}\matr{Q}(t),
		\label{Atilde}
	\end{equation}
	where $\matr{Q}(t) = \text{diag}(n_1(t),...,n_N(t))$, with $n_i(t)=1$ if node (column) $i$ is scheduled (sampled) at iteration $t$. Note that we can schedule multiple collision-free subsets in every round; therefore, the scheduling decision is made on the subsets instead of the nodes directly.
	A visual example is given in Fig. \ref{Link_removal}(c)-(d).
	Using (\ref{Atilde}), the Laplacian matrix of the effective communication topology in iteration $t$ is given as
	\begin{equation}
		\tilde{\matr{L}}(t)=\text{diag}(\tilde{\matr{A}}(t)\vect{u})-\tilde{\matr{A}}(t),
	\end{equation} 
	which is symmetric, and each row/column of the matrix sums up to zero.
	The effective mixing matrix in iteration $t$ is then defined as
	\begin{equation}
		\matr{W}(t) = \matr{I}-\epsilon\tilde{\matr{L}}(t),
		\label{Wt_symm}
	\end{equation}
	which is also symmetric, and each row/column of the matrix sums up to one. Note that $\epsilon$ is constant and independent of $t$.
	
	\subsection{Designing Scheduling Probabilities and Mixing Matrix}
	\label{mixing_opt}
	
	\begin{figure*}[ht!]	
		\begin{equation}
			\left(\mathbb{E}[\tilde{\matr{L}}(t)]\right)_{ij} =
			\begin{cases}
				-p_i\max\{\phi(i,j),p_j\} (\matr{A})_{ij}, & \text{if } i \neq j \\
				\sum_{m=1}^{N} p_i\max\{\phi(i,m),p_m\}(\matr{A})_{im}(\matr{A})_{mi}, & \text{if } i = j
			\end{cases}
			\label{ELt}
		\end{equation}
		
		\begin{equation}			
			\mathbb{E}\left[\tilde{\matr{L}}^T(t)\tilde{\matr{L}}(t)\right]  = \mathbb{E}\left[\text{diag}^2(\tilde{\matr{A}}\vect{u})\right] - \mathbb{E}\left[\text{diag}(\tilde{\matr{A}}\vect{u})\tilde{\matr{A}}\right] - \mathbb{E}\left[\tilde{\matr{A}}\text{diag}(\tilde{\matr{A}}\vect{u})\right] + \mathbb{E}\left[\tilde{\matr{A}}^2 \right]
			\label{EL2}
		\end{equation}	
		
		\begin{equation}
			\left(\mathbb{E}\left[ \text{diag}^2(\tilde{\matr{A}}\vect{u})  \right]\right)_{ij} =
			\begin{cases}
				0, & \text{if } i \neq j \\
				\sum_{m=1}^{N}\sum_{k=1}^{N} p_i\max\{\phi(i,m),p_m\}\max\{\phi(i,k),\phi(m,k),p_k\}
				(\matr{A})_{im}(\matr{A})_{ik}, & \text{if } i = j
			\end{cases}
		\end{equation}	
		
		\begin{equation}
			\left(\mathbb{E}\left[ \text{diag}(\tilde{\matr{A}}\vect{u})\tilde{\matr{A}}  \right]\right)_{ij} =
			\begin{cases}
				\sum_{m=0}^{N}p_i\max\{\phi(i,j),p_j\}\max\{\phi(i,m),\phi(j,m),p_m\}
				(\matr{A})_{ij}(\matr{A})_{im}, & \text{if } i \neq j \\
				0, & \text{if } i = j
			\end{cases}	
		\end{equation}		
		
		\begin{equation}
			\left(\mathbb{E}\left[ \tilde{\matr{A}}\text{diag}(\tilde{\matr{A}}\vect{u})  \right]\right)_{ij} =
			\begin{cases}
				\sum_{m=0}^{N}p_i\max\{\phi(i,j),p_j\}\max\{\phi(i,m),\phi(j,m),p_m\}
				(\matr{A})_{ij}(\matr{A})_{jm}, & \text{if } i \neq j \\
				0, & \text{if } i = j
			\end{cases}	
		\end{equation}	
		
		\begin{equation}
			\left(\mathbb{E}\left[ \tilde{\matr{A}}^2\right]\right)_{ij} =
			\begin{cases}
				\sum_{m=0}^{N}p_i\max\{\phi(i,j),p_j\}\max\{\phi(i,m),\phi(j,m),p_m\}
				(\matr{A})_{im}(\matr{A})_{mj}, & \text{if } i \neq j \\
				\sum_{m=0}^{N}p_i\max\{\phi(i,m),p_m\}
				(\matr{A})_{im}^2, &  \text{if } i = j
			\end{cases}	
			\label{last_eq}
		\end{equation}	
		\hrulefill	
	\end{figure*}
	
	As mentioned earlier, not all nodes are equally important in the network, and more important nodes should be scheduled more often. 
	Recall that in every iteration, subset $\mathcal{S}_k$ is scheduled with probability $p_{\mathcal{S}_k}$, for $k\in\{1,\ldots,q\}$. Then we have
	\begin{equation}
		\mathbb{E}[\matr{Q}(t)]=\text{diag}(p_1,...,p_N),
	\end{equation}
	where the activation probability of a node equals the scheduling probability of the subset to which it belongs, i.e. $p_i = p_{\mathcal{S}_k}$ if $i\in\mathcal{V}_k$.
	Therefore, the number of scheduled subsets in each iteration is a random number, and on average, the number of required transmission slots per iteration is $\sum_{l=1}^q p_{\mathcal{S}_l}$. 
	
	For accelerating per-round convergence with constrained communication, the scheduling probabilities should be optimized by solving the following optimization problem:
	\begin{equation}
		\begin{aligned}
			\min_{p_{\mathcal{S}_1},...,p_{\mathcal{S}_q}} \quad & ||\mathbb{E}[\matr{W}^2(t)]-\matr{J}||_2 \\
			\textrm{s.t.} \quad & \sum_{l=1}^q p_{\mathcal{S}_l} = \mathcal{B}, \\
			\quad & 0\leq p_{\mathcal{S}_m} \leq 1, \forall m. \\
		\end{aligned}
		\label{P_colors optimization}
	\end{equation}
	Here, $\mathcal{B}$ is the communication budget, which indicates the average number of transmission slots (also the average number of activated subsets) per iteration. The expression for $\mathbb{E}[\matr{W}^2(t)]$ includes several nonlinear terms given by products of the probabilities, which makes the optimization problem very challenging. 
	For this reason, we adopt a heuristic approach for designing the scheduling probabilities, based on the idea presented in \cite{herrera2023distributed}. Each node $i$ has a betweenness centrality value $b_i$ and $\sum_{i=1}^{N}b_i=1$. We can define the betweenness centrality of a subset as 
	\begin{equation}
		b_{\mathcal{S}_j} = \sum_{i=1}^{N} b_i \mathbf{1}_{\{i\in\mathcal{V}_j\}}
	\end{equation}
	where $\mathbf{1}_{\{i\in\mathcal{V}_j\}}$ is an indicator function, which is equal to one if $i\in\mathcal{V}_j$, and zero otherwise.
	Finally, we choose
	\begin{equation}
		p_{\mathcal{S}_j} = \min\{1,\gamma b_{\mathcal{S}_j}\}
	\end{equation}
	where the constant $\gamma$ is chosen such that $\sum_{i=1}^{q}p_{\mathcal{S}_i}=\mathcal{B}$.
	
	Another key factor to speed up convergence of D-SGD is the design of the mixing matrix.
	Choosing $\matr{W}(t)$ in every round as in (\ref{Wt_symm}), for given scheduling probabilities, allows us to optimize the mixing matrix by solving problem (\ref{P_colors optimization}) with optimization variable $\epsilon$ instead. Using similar techniques as in \cite{wang2022matcha}, this problem can be reformulated as follows:
	\begin{equation}
		\begin{aligned}
			\min_{s, \epsilon, \beta} \quad & s \\
			\textrm{s.t.} \quad & \epsilon^2-\beta\leq 0,\\
			& \matr{I} -2\epsilon\mathbb{E}\left[\tilde{\matr{L}}(t)\right] + \beta\left(\mathbb{E}\left[\tilde{\matr{L}}^T(t)\tilde{\matr{L}}(t)\right]\right) -\matr{J}\preccurlyeq s\matr{I}, \\
		\end{aligned}
		\label{epsilon_opt}
	\end{equation}
	which is a convex problem, where $s$ and $\beta$ are auxiliary variables.
	If we define a function $\phi:\mathbb{R}^2\rightarrow\mathbb{R}$, with $\phi(i,j)=1$ if nodes $i$ and $j$ are in the same subset, and zero otherwise, we can express the $(i,j)$-th element of $\mathbb{E}\left[\tilde{\matr{L}}(t)\right]$ as in (\ref{ELt}). Note that $\max\{\phi(i,j),p_j\}=p_j$ if nodes $i$ and $j$ are in different subsets and $\max\{\phi(i,j),p_j\}=1$ if not. 
	Similarly, we can express the $(i,j)$-th element of $\mathbb{E}\left[\tilde{\matr{L}}^T(t)\tilde{\matr{L}}(t)\right]$ as in (\ref{EL2})-(\ref{last_eq}).
	
	\subsection{Discussions on Comparison with Existing Approaches}	
	\label{comparison_and_discussion}
	There are several existing works considering link-based scheduling for accelerating D-SGD with limited communication cost, such as MATCHA in \cite{wang2022matcha}. However, MATCHA is not directly applicable in the wireless setting, because the partition rule does not satisfy the collision-free condition in wireless networks. 
	In addition, even if the partition rule is modified, one matching will still require two transmission slots to achieve bi-directional communication. In some topologies with densely connected local parts, such as the examples shown in Fig. \ref{exp_fig}, broadcast-based scheduling can activate many more links as compared to link-based scheduling under the same communication budget. This is the main motivation behind our work. 
	
	\section{Numerical Experiments}
	
	\begin{figure*}[ht!]	
		\begin{subfigure}[c]{0.3\linewidth}
			\includegraphics[width=\linewidth]{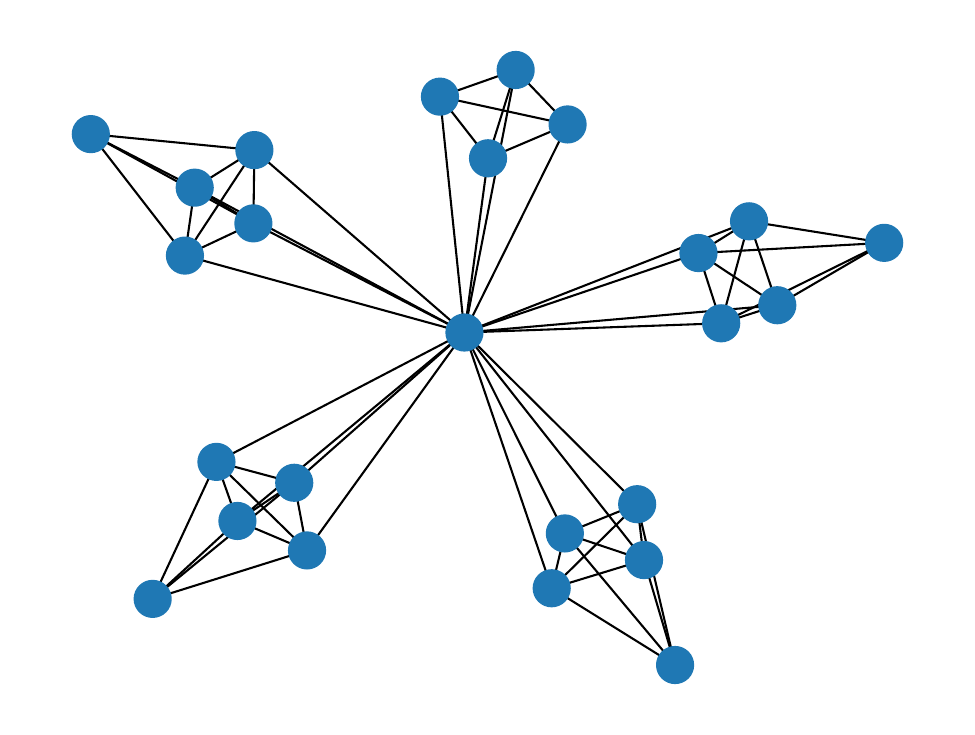}
		\end{subfigure}
		\hfill		
		\begin{subfigure}[c]{0.3\linewidth}
			\includegraphics[width=\columnwidth]{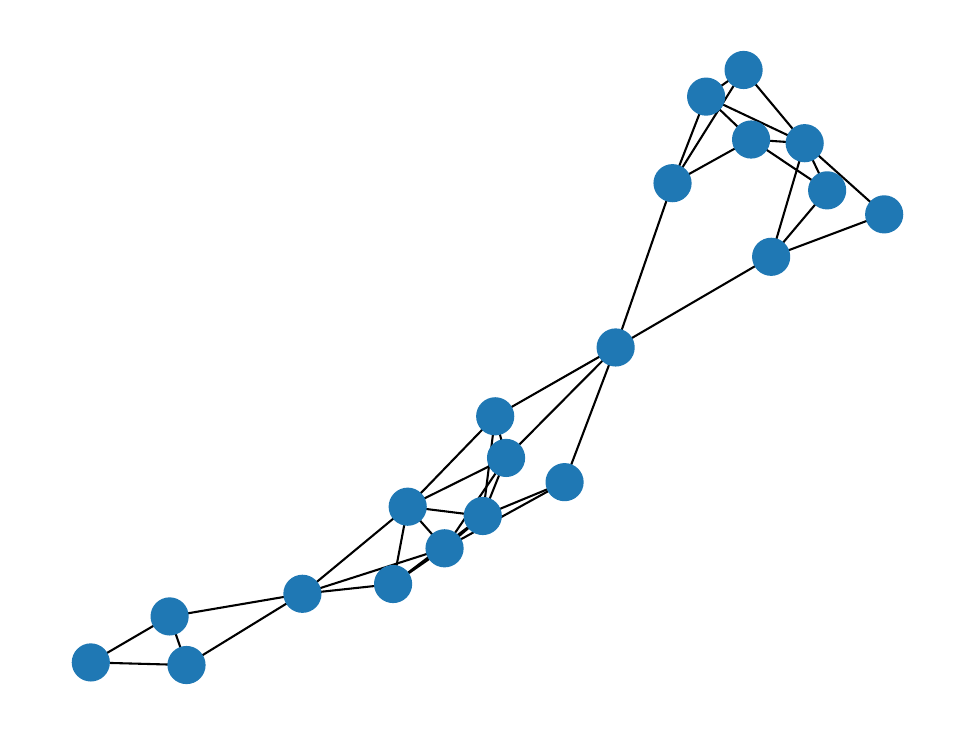}
		\end{subfigure}
		\hfill
		\begin{subfigure}[c]{0.3\linewidth}
			\includegraphics[width=\linewidth]{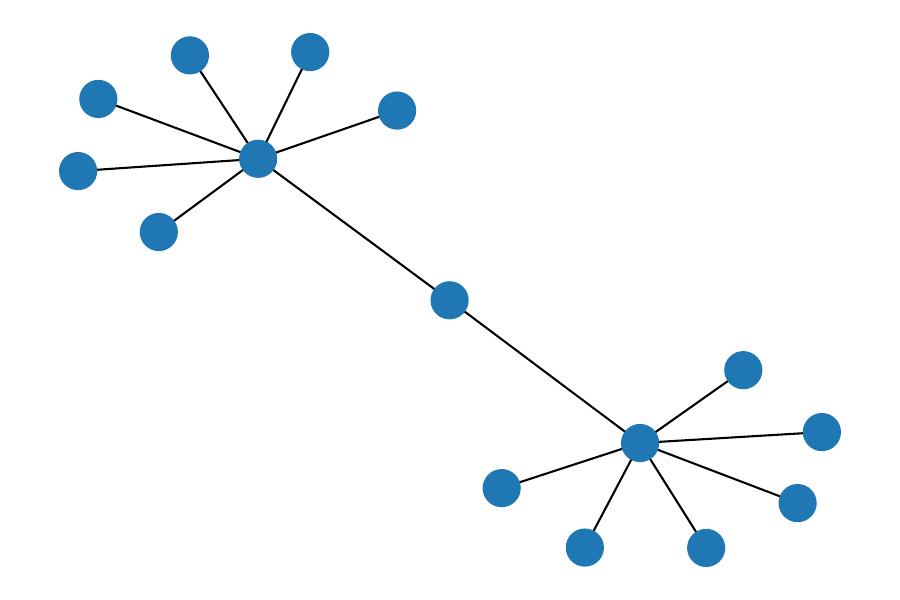}
		\end{subfigure}
		
		\begin{subfigure}[c]{0.3\linewidth}
			\includegraphics[width=\linewidth]{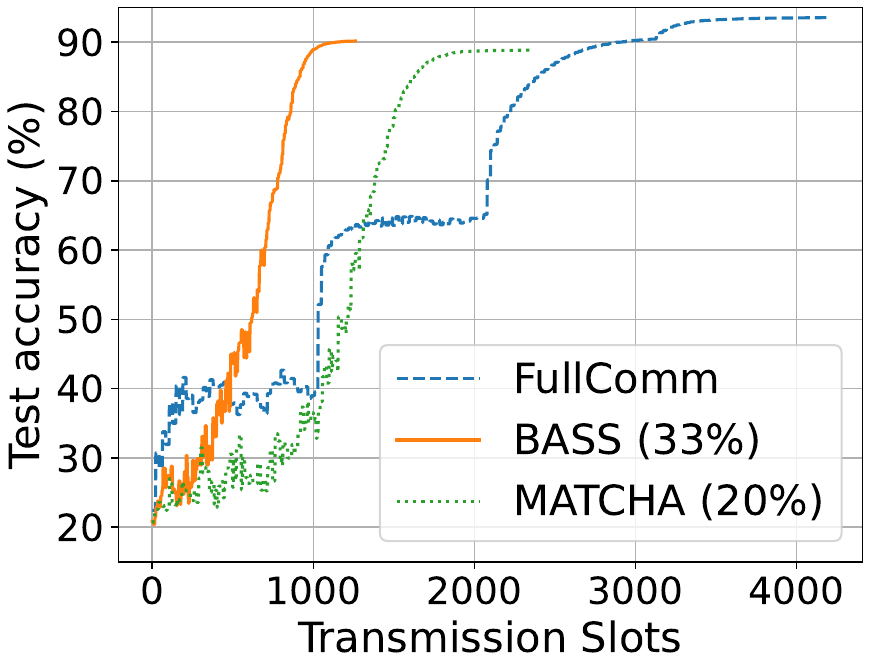}
		\end{subfigure}
		\hfill
		\begin{subfigure}[c]{0.3\linewidth}
			\includegraphics[width=\columnwidth]{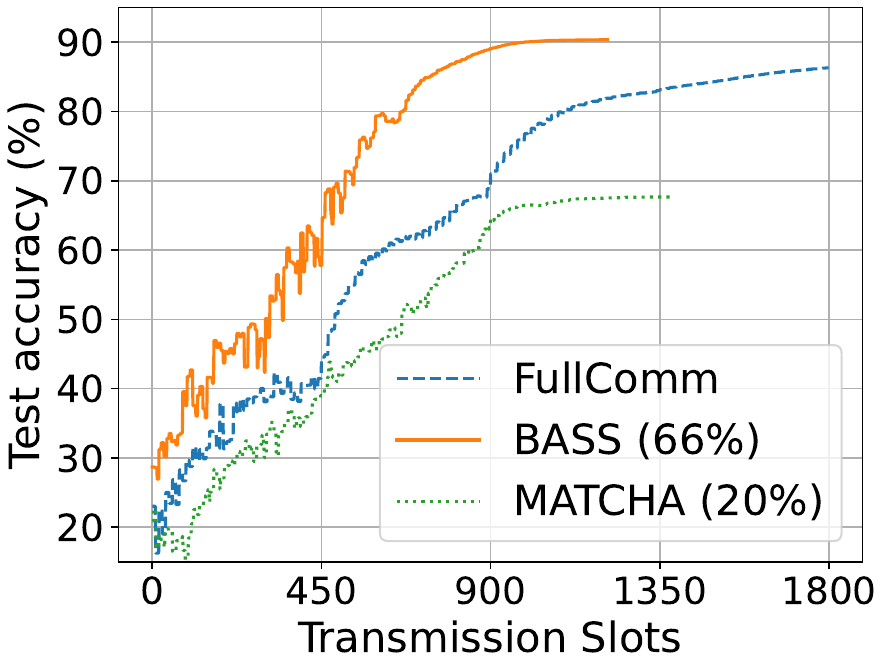}
		\end{subfigure}
		\hfill
		\begin{subfigure}[c]{0.3\linewidth}
			\includegraphics[width=\linewidth]{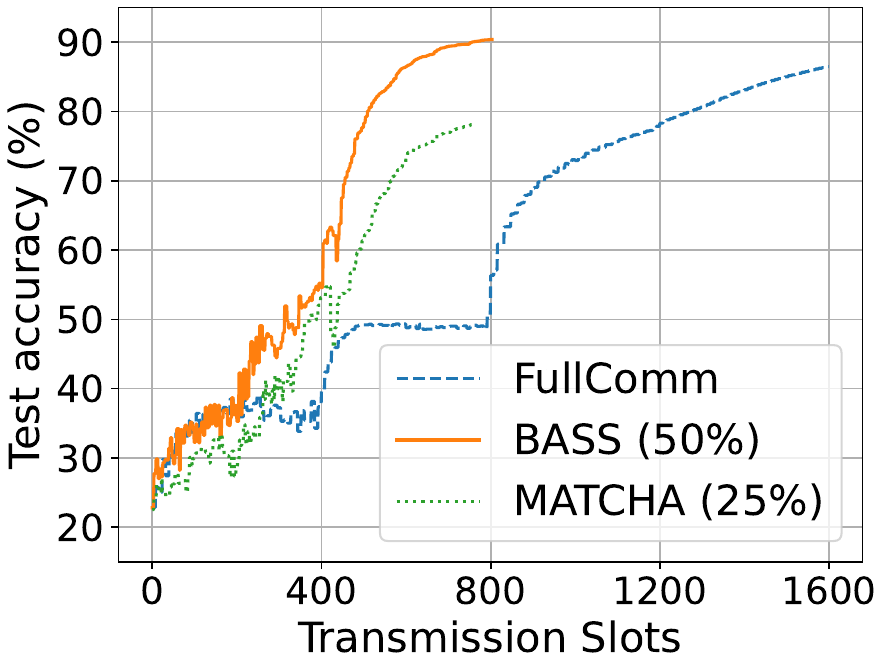}
		\end{subfigure}
				
		\caption{Performance comparison between BASS, modified MATCHA, and full communication in different topologies.}
		\label{exp_fig}
		
	\end{figure*}
	
	In this section, we evaluate the performance of our proposed broadcast-based subgraph sampling method for D-SGD over wireless networks. We consider three network topologies as illustrated in Fig. \ref{exp_fig}, with $N=\{25,20,15\}$ nodes respectively. Simulations are performed by training a multi-layer perceptron (MLP) for an image classification task. 
	We use the MNIST dataset, which contains $60000$ images for
	training and $10000$ for testing. 
	The training dataset was divided into $2N$ shards (for a total of $N$ users), and each user's local dataset is given by $2$ randomly selected shards with no repetition.
	We use \textit{SparseCategoricalCrossentropy} from Keras as loss function and ADAM optimizer with 0.01 weight decay and a decaying learning rate. Each experiment runs for $200$ communication rounds (iterations).
	The performance of our proposed design is evaluated by showing the improvement of test accuracy with the number of transmission slots.
	For performance comparison, we also implement a modified version of MATCHA, and the full communication case, where all nodes are activated in every round.  
		
	In Fig. \ref{exp_fig}, we compare the performance of BASS, modified MATCHA and full communication for different topologies. 
	Note that full communication is a special case of BASS if all subsets are scheduled in every round. We refer to this case as either ``FullComm'' or ``BASS(100\%)''.
	The percentage next to MATCHA and BASS represents the fraction of the total number of matchings and subsets that are activated on average per round, respectively. For the experiments with BASS and MATCHA, we have tried different activation percentages, and the one that gives the best performance is plotted in the figure.
	We observe that BASS clearly outperforms modified MATCHA by a considerable margin, since the former takes more advantage of broadcast transmissions as compared to the latter, and this allows it to use less transmission slots to activate more links and transmit more information.
	Another observation is that MATCHA does not always outperform the full communication case, indicating that partial communication with link scheduling is not always beneficial for improving convergence per transmission slot. 
	
	\begin{figure}[t!]
		\centering
		\includegraphics[width=0.85\columnwidth]{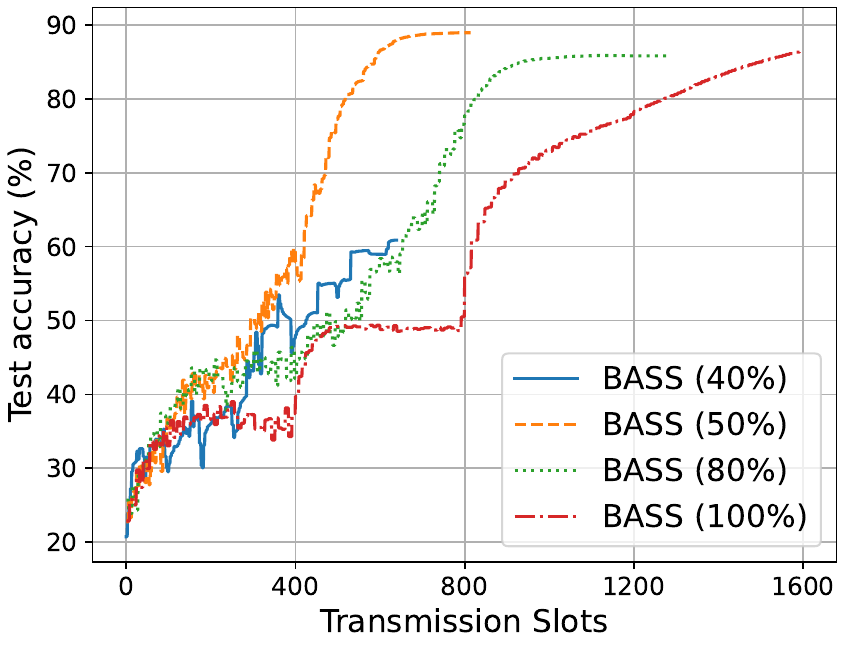}
		\caption{Impact of the communication budget.}
		\label{comm_bud_comparison}
	\end{figure}
	
	Finally, in Fig. \ref{comm_bud_comparison}, we show the impact of communication budget per iteration on the learning performance. Here, we only consider BASS, using the third graph in Fig. \ref{exp_fig} (two connected stars) as the base topology. 
	From this figure, we observe that the effect of partial communication varies a lot depending on the communication budget per iteration. On one hand, if the communication budget is very low (small activation percentage), the number of activated links per iteration is very low, which leads to poor information fusion/mixing. On the other hand, if the budget is very high, the communication improvement per transmission slot becomes marginal, leading to inefficient use of resources. 
	Optimal allocation of communication budget per iteration remains to be investigated in future work.
				
	\section{Conclusion}

	In this work, we propose BASS, a broadcast-based subgraph sampling method for accelerating decentralized learning over wireless networks with partial communication. 
	The base topology is divided into multiple collision-free subsets, where in each subset all nodes can simultaneously broadcast model updates to their neighbors. 
	We adopt a probabilistic approach to activate these subsets by considering the importance of the nodes on the connectivity of the graph. 
	Simulation results show that BASS significantly outperforms existing link-based scheduling policies and plain D-SGD, by achieving faster convergence with fewer number of transmission slots.
	
	The essence behind this work is similar to MATCHA \cite{wang2022matcha}, motivated by the fact that connectivity-critical subgraphs should be activated more often, when considering the actual communication cost or delay per iteration. 
	With this work, we further demonstrate that broadcast-based communication can be exploited to accelerate convergence (measured by improvement per transmission slot) of D-SGD over wireless networks.


\end{document}